# Fuzzy Modeling and Natural Language Processing for Panini's Sanskrit Grammar

P. Venkata Subba Reddy

**Abstract**—Indian languages have long history in World Natural languages. Panini was the first to define Grammar for Sanskrit language with about 4000 rules in fifth century. These rules contain uncertainty information. It is not possible to Computer processing of Sanskrit language with uncertain information. In this paper, fuzzy logic and fuzzy reasoning are proposed to deal to eliminate uncertain information for reasoning with Sanskrit grammar. The Sanskrit language processing is also discussed in this paper.
.

**Index Terms**— Fuzzy logic, Fuzzy reasoning, Natural language processing, Panini's Sanskrit grammar

——————————— ◆ ———————————

## 1 INTRODUCTION

Computer languages are defined based on particular grammar, for example C, Pascal etc. Computer processing of natural languages has become an application area such as natural language processing, Machine translation etc.Natural languages are to be defined with the grammer.

Panini defined Sanskrit grammar with 4000 rules long back in sixth century B.C. [4]. The rules of Panini's Sanskrit grammar contain uncertainty. Computer processing of this grammar is very difficult unless uncertainty is eliminated. There are many logics available to deal with uncertainty like fuzzy logic, probabilistic logic etc.

Fuzzy logic and reasoning [7, 8, 13, 14] are discussed to deal with uncertainty in the rules of Panini's Sanskrit grammar. The Paninian approach to Natural Language Processing (NLP) is reviewed and compared with the current computer-based understanding systems [6, 12, 15]. The Paninian-style generative rules and meta-rules could assist in further advances in NLP.

## 2 FUZZY LOGIC

Zadeh[9] proposed Fuzzy logic to deal with Uncertain and Incomplete information. Here we discuss Fuzzy logic briefly and later we apply it for Panini's Sanskrit Grammar [2,3].

Zadeh[9] has introduced Fuzzy set as a model to deal with imprecise, inconsistent and inexact information. Fuzzy set is a class of objects with a continuum of grade of membership.

The Fuzzy set A of X is characterized by its membership function $A = \mu_A(x)$ and ranging values in the unit interval $[0, 1]$

$\mu_A(x): X \rightarrow [0, 1]$, $x \in X$, where X is Universe of discourse.
or
$A = \mu_A(x_1)/x_1 + \mu_A(x_2)/x_2 + \ldots + \mu_A(x_n)/x_n$

Where "+" is union
For instance, x is Tall is defined as

Tall = $\mu_{Tall}(x) \rightarrow [0, 1]$, where "Tall" is fuzzy set.

Tall = $0.56/x_1 + 0.6/x_2 + 0.65/x_3 + 0.67/x_4 + 0.69/x_5$

The combination of Fuzzy sets are given as follows

$A \land B = \min\{\mu_A(x), \mu_B(x)\}/x$ (disjunction)
$A \lor B = \max\{\mu_A(x), \mu_B(x)\}/x$ (conjunction)
$A' = 1 - \mu_A(x)/x$ (negation)
$A \rightarrow B = \max\{1 - \mu_A(x), \mu_B(x)\}/x$ (implication)
$\mu_A(x, y) = \{\mu_A(x) \times \mu_B(y)\}/x, y = \min\{\mu_A(x) \times \mu_B(y)\}/x, y$
$A \circ R = \max\{\mu_A(x), \mu_A(x, y)\}/x$ where $R(x, y)$ is Fuzzy relation and "o" is composition

The propositions may contain quantifiers. Square operator is used for 'very', 'most' (concentration). The square root operator is used for 'more or less' (diffusion).

For instance,

$\mu_{very\ young}(x) = \mu_{young}(x)^2$
$\mu_{notvery\ young}(x) = 1 - \mu_{young}(x)^2$
$\mu_{more\ or\ less\ young}(x) = \mu_{young}(x)^{1/2}$

Fuzzy reasoning[8] is a drawing conclusion from Fuzzy propositions using fuzzy inference rules. Some of the Fuzzy inference rules are given bellow

R1:  x is A           R2: x is A
     x and y are B        x or y is B
     ―――――――――            ―――――――――
     y is A∧B             y is A∨B

R3:  x and y are A    R4: x or y are A
     y and z are B        y or z are B
     ―――――――――            ―――――――――
     y and z are B        x or z are B

R5: x is A
if x is A then y is B
_______________
y is A o (A→B)

## 3  FUZZY MODELLING FOR PANINI'S SANSKRIT GRAMMAR

Grammars are defined to either programming languages or natural languages. Computer processing of natural languages and language translations is an application area in the computer field. Indian languages are having long history. Panini proposed grammar with 4000 rules for Sanskrit. These are categorized into different sets. One of them is Syadvada set. The Syadvada set contains seven possibilities they are given below.

1. May be, it is. (Syadasti)
2. May be, it is not (Sada nasti)
3. May be it is, and it is not at different times (Syad asti-nasti)
4. May be it is and it is not at the same time and is indescribable (Syad avaktavya)
5. May be it is, and yet indescribable (Syad asti avaktavya)
6. May be it is not, and also indescribable (Syad asti nasti avaktavya)
7. May be it is, and it is not and also indescribable ( Syad asti nasti avaktavya)

The above rules contain uncertainty. The uncertainty has to be eliminated for further computer processing of the Sanskrit language. Fuzzy logic can be used to eliminate the uncertainty. The fuzzy logic is applied for the above rules to eliminate uncertainty and the rules are given below.

1. May be, it is. ( Syadasti)
   $\mu_{Syadasti}(x) \rightarrow [0,1]$

2. May be, it is not (Syad nasti)
   Syad nasti = $1 - \mu_{Syadasti}(x)$

3. May be it is, and it is not at different times (Syad asti-nasti)

   $\mu_{Syadasti}(x) \wedge (1 - \mu_{Syadasti}(x)$ o $\mu_{different\ times}(x,y))$
   where "o" is composition and "y" is time variable

4. May be it is and it is not at the same time and is indescribable

   $(\mu_{Syadasti}(x) \wedge (1 - \mu_{Syadasti}(x)$ o $\mu_{different\ times}(x,t)) \wedge \mu_{different\ times}(x)$
   where "t" is constant.

5. May be it is and yet indescribable. (Syad asti avaktavya)

   $\mu_{Syadasti}(x) \wedge \mu_{different\ times}(x)^{1/2}$
   where yet is diffusion

6. May be it is not, and also indescribable (yad asti nasti avaktavya)
   $(1 - \mu_{Syadasti}(x)) \wedge \mu_{different\ times}(x)$

7. May be it is, and it is not and also indescribable (Syad asti nasti avaktavya)
   $\mu_{Syadasti}(x) \wedge (1 - \mu_{Syadasti}(x) \wedge \mu_{different\ times}(x))$

This fuzzy representation of the sanskrit sentences shall be further used for fuzzy reasoning [7, 13, 14, 16].
For istance, consider two sentences

May be, it is. (Syadasti)
May be it is, and it is not at different times (Syad asti-nasti)

The inference will be given as using R1

"it is not at different times " with the fuzziness
(Syadasti) ∧ (Syad asti-nasti)

## 4  SANSKRIT LANGUAGE PROCESSING

Beattie[1]  presents an introductory review of some aspects of the computer processing of natural language  in the form of a string of alphabetic characters, for example, spoken word. Applications of such processing in fields like information storage and retrieval and computer-assisted instruction are discussed for a computer to "understand" natural language[3].

The Sanskrit Language can be processed by defining English alphabetic characters using OM SETUP. This Sanskrit language representation in English shall be used for reasoning with the Sankrit language

Zenon [16] describes a new programming language FLISP which provides a number of facilities for efficiently representing and manipulating fuzzy knowledge. It is based on fuzzy sets and fuzzy logic theories. The language contains a collection of fuzzy-set operations and procedures for solution of fuzzy relational equations with triangular norms. FLISP forms implement and examine the fuzzy control algorithms. FLISP may be used to deal with Uncertainty in Panini's Sanskrit Grammer.

PROLOG can also be used for reasoning with the Sanskrit language.

## 5 COCLUSION

Panini's proposed Sanskrit Grammar to Sanskrit language. The rules contain uncertainty information. The elimination of uncertainty information with Fuzzy logic is discussed. Fuzzy logic and fuzzy reasoning are discussed to deal with uncertainty information in Panini's Sanskrit Grammar to make it convenient for further computer processing. The Computer processing of Sanskrit language is discussed.

## AKNOWLEDGEMENT

Our special thanks to Prof. V.V.S.Sarma, Computer Science and Automation, Indian Institute of Science, Bangalore for discussion, and reviewers for their valuable suggestions.

**P. Venkata Subba Reddy** is working as Associate Professor in Department of Computer science and Engineering, College of Engineering, Sri Venkateswara University, Tirpathi, India since 2001. He joined as Asst. Professor in Department of Computer science and Engineering, College of Engineering, Sri Venkateswara University in 1992 and promoted as Associate Professor in 2001. He did Post Graduate degree in Applied Mathematics with Computer Programming as Specilization during 1984-86. He did his Post Graduation Diploma in Computer Methods & Programming from Computer Society of India, Hyderabad. He did M.Phii in Database Management systems during 1986-88 and Ph.D in Artificial Intelligence during 1988-1992 in Sri Venkateswara University, Tirpathi, India. . He did Post Doctoral/Visiting fellowship in Fuzzy Algorithms from IISC/JNCAR, Bangalore, India under Prof. V. Rajaraman in 1996. He is actively engaged in Teaching and Research work to B.Tech., M.Tech., and Ph.D students. He published papers in reputed journals. He is an Editor for JCSE.